\documentclass[letterpaper, 10 pt, conference]{ieeeconf}  

\IEEEoverridecommandlockouts                              

\usepackage{graphicx}
\usepackage{caption}
\usepackage{subcaption}
\usepackage{xspace}

\newcommand{\nav}{\textbf{Navigation}\xspace}
\newcommand{\manip}{\textbf{Manipulation}\xspace}

\title{\LARGE \bf
Evaluating Customization of Remote Tele-operation Interfaces for Assistive Robots
}

\author{Vinitha Ranganeni, Noah Ponto and Maya Cakmak
\thanks{All authors are associated with Paul G. Allen School of Computer Science \& Engineering, University of Washington,
        \{vinitha, ponton, mcakmak\}@cs.washington.edu}%
}

\begin{document}

\maketitle
\thispagestyle{empty}
\pagestyle{empty}

\begin{abstract}

Mobile manipulator platforms, like the Stretch RE1 robot, make the promise of in-home robotic assistance feasible. For people with severe physical limitations, like those with quadriplegia, the ability to tele-operate these robots themselves means that they can perform physical tasks they cannot otherwise do themselves, thereby increasing their level of independence. In order for users with physical limitations to operate these robots, their interfaces must be accessible and cater to the specific needs of all users. As physical limitations vary amongst users, it is difficult to make a single interface that will accommodate all users. Instead, such interfaces should be customizable to each individual user. In this paper we explore the value of customization of a browser-based interface for tele-operating the Stretch RE1 robot. More specifically, we evaluate the usability and effectiveness of a customized interface in comparison to the default interface configurations from prior work. We present a user study involving participants with motor impairments (N=10) and without motor impairments, who could serve as a caregiver, (N=13) that use the robot to perform mobile manipulation tasks in a real kitchen environment. Our study demonstrates that no single interface configuration satisfies all users' needs and preferences. Users perform better when using the customized interface for navigation, but not for manipulation due to higher complexity of learning to manipulate through the robot. All participants are able to use the robot to complete all tasks and participants with motor impairments believe that having the robot in their home would make them more independent.

\end{abstract}

\section{INTRODUCTION}
Physically assistive robots have the potential to assist people with motor limitations to complete activities of daily living independently. However, these robots do not yet have robust autonomous capabilities for completing tasks in a wide variety of environments. 
Tele-operation can make these robots more readily available and satisfy users' desire for having control.

\begin{figure}
\centering
  \includegraphics[width=\columnwidth]{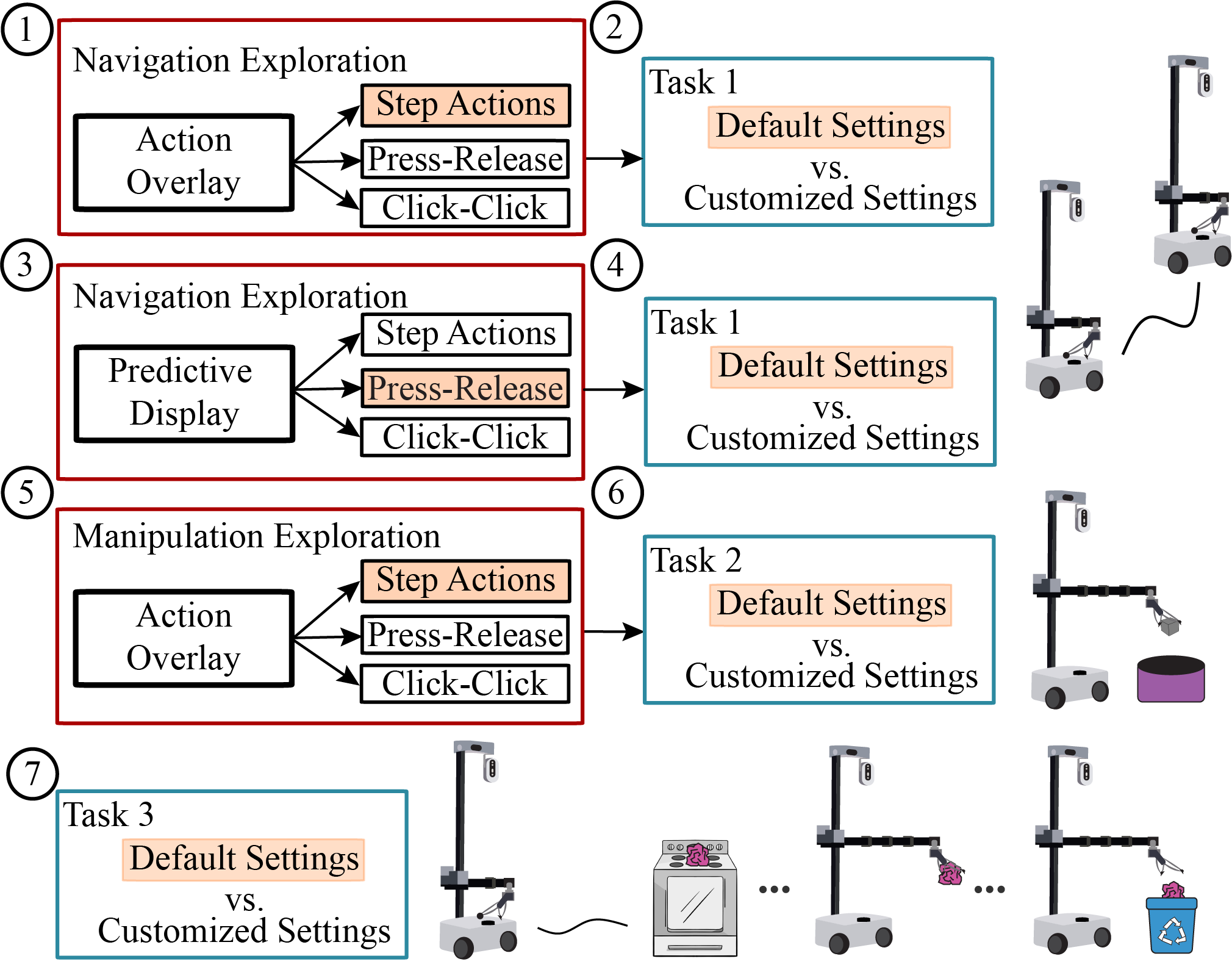}
  \caption{An overview of the study design. Users go through three exploration phases to learn how to use the various control display modes (action overlay and predictive display)  and action modes (step actions, press-release, and click-click) in the interface. After each exploration phase they complete a task with their customized settings and the default settings that are highlighted in orange.}
  \label{fig:study_overview}
\end{figure}

Many existing tele-operation interfaces provide a single control configuration which may not be accessible to all users. In this work we explore customization of remote tele-operation interfaces for operating a Stretch RE1. More specifically, we build on prior work done by Cabrera et. al~\cite{cabrera2021exploration} by adding additional control features and analyzing user preferences and performance when using different interface configurations. We run two studies with users without motor impairments, who could serve as a caregiver, (N=13) and users with motor impairments (N=10). In these studies (Fig.~\ref{fig:study_overview}) users learn how to use the various control settings in the interface and are asked to complete a series of tasks using default settings determined by prior work and their own customized settings. We have three hypotheses:
\begin{itemize}
    \item[\textbf{H1}] The is no single interface configuration that will satisfy all users' needs and preferences.
    \item[\textbf{H2}] Users with motor impairments will have different preferences than people without.
    \item[\textbf{H3}] Users' task completion time, number of errors and clicks will be lower when using their customized settings.
\end{itemize}

Our findings show that users' preferences in interface configurations vary and there is no single configuration that is the ``winner". All users perform better when using their customized settings for the control of navigation. When controlling manipulation, they perform better when completing the task the second, time irrespective of interface configuration, likely due to the complexity of manipulation through the robot. 
Additionally, users found the interface to be intuitive, easy-to-learn, easy-to-use in all configurations and found the robot useful. Participants with motor impairments believe that having the robot in their home would make them more independent, demonstrating the utility of tele-operated assistive robots in the near future.

\begin{figure}
\centering
  \includegraphics[width=\columnwidth]{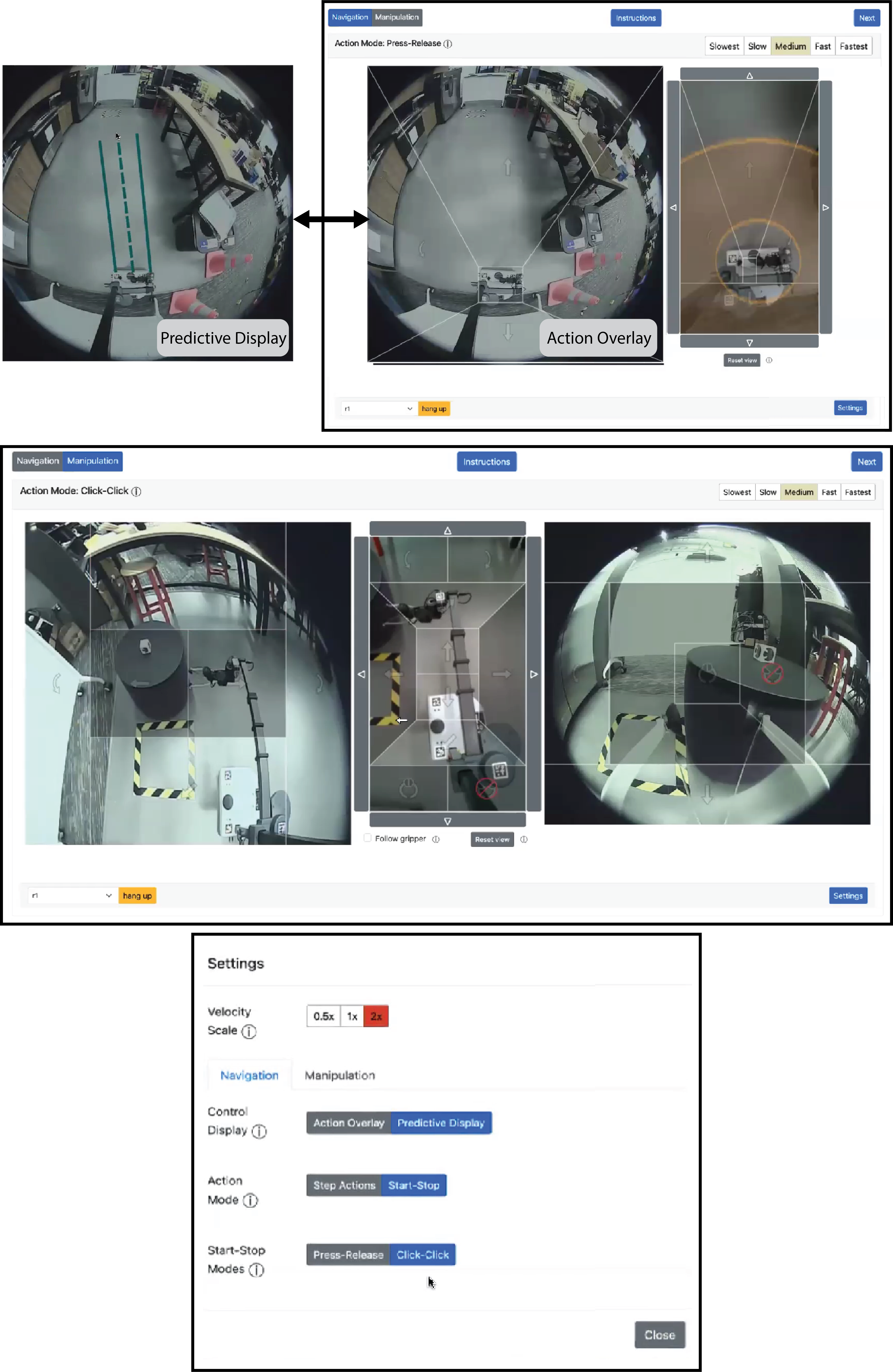}
  \caption{(Top) Interface in \textbf{Navigation} mode. There are two possible control displays: action overlay and predictive display. (Middle) The interface in \textbf{Manipulation} mode. (Bottom) The settings menu.}
  \label{fig:interface}
\end{figure}

\section{Related Works}
Prior work has explored the potential for assistive robots to assist people with motor impairments~\cite{chen2013robots, brose2010role}. Tele-operation of these robots has been shown to be a viable solution to enable individuals with motor limitations to complete activities of daily living (ADL) independently~\cite{ciocarlie2012mobile, grice2019home, park2020active}. Additionally, tele-operation allows robots to be practical without requiring full autonomy and satisfies the users' desire to have control. 

 To reduce the burden on the user while still giving them control, semi-autonomous tele-operation systems have been widely studied. A significant amount of this work studies inferring user intent during tele-operation and providing autonomous assistance accordingly~\cite{dragan2013teleoperation, hauser2013recognition, khokar2014novel, gopinath2021customized, herlant2016assistive, admoni2016predicting}. However, these tele-operation interfaces provide a single control configuration that may not be accessible to users with different abilities or necessarily satisfy users' preferences. Making control interfaces customizable will make them accessible to users with unique physical abilities and fit user preferences. A participant, with motor impairments, in a user study from prior work specifically emphasized the need for flexible interfaces that cater to people with motor impairments~\cite{cabrera2021exploration}. Furthermore, prior work has shown that allowing people to customize tele-operation interfaces impacts their task completion time and subjective preferences~\cite{cabrera2021cursor}. 

Much of the work on interface customization has been for GUIs~\cite{hurst2008automatically, carter2006dynamically, gajos2007automatically}. Jain et. al have explored customization of the level of assistance of provided by the robot~\cite{jain2016approach} but not the control interface itself. In this work, we explore customization of cursor-based web interface for remotely tele-operating a mobile manipulator. We build on work done by Cabrera et. al~\cite{cabrera2021exploration} who developed a web interface for remotely tele-operating a Stretch RE1. We developed additional control features for tele-operating Stretch and analyze subjective preferences, task performance, and success.

\begin{figure*}
     \centering
     \begin{subfigure}[b]{0.285\textwidth}
         \centering
         \includegraphics[width=\textwidth]{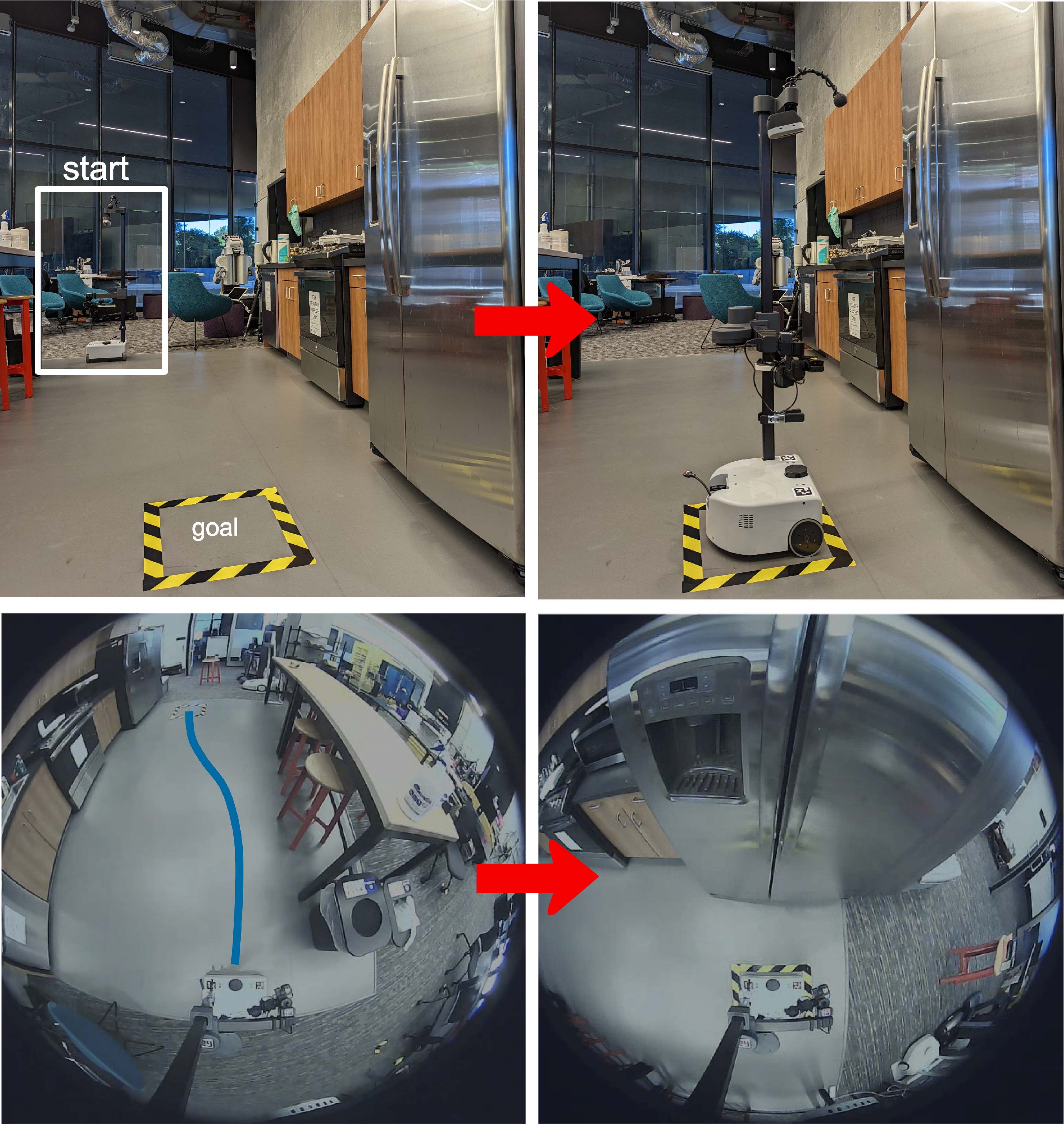}
         \caption{Task 1:  Navigate to the fridge}
         \label{fig:task_1}
     \end{subfigure}
     \hfill
     \begin{subfigure}[b]{0.285\textwidth}
         \centering
         \includegraphics[width=\textwidth]{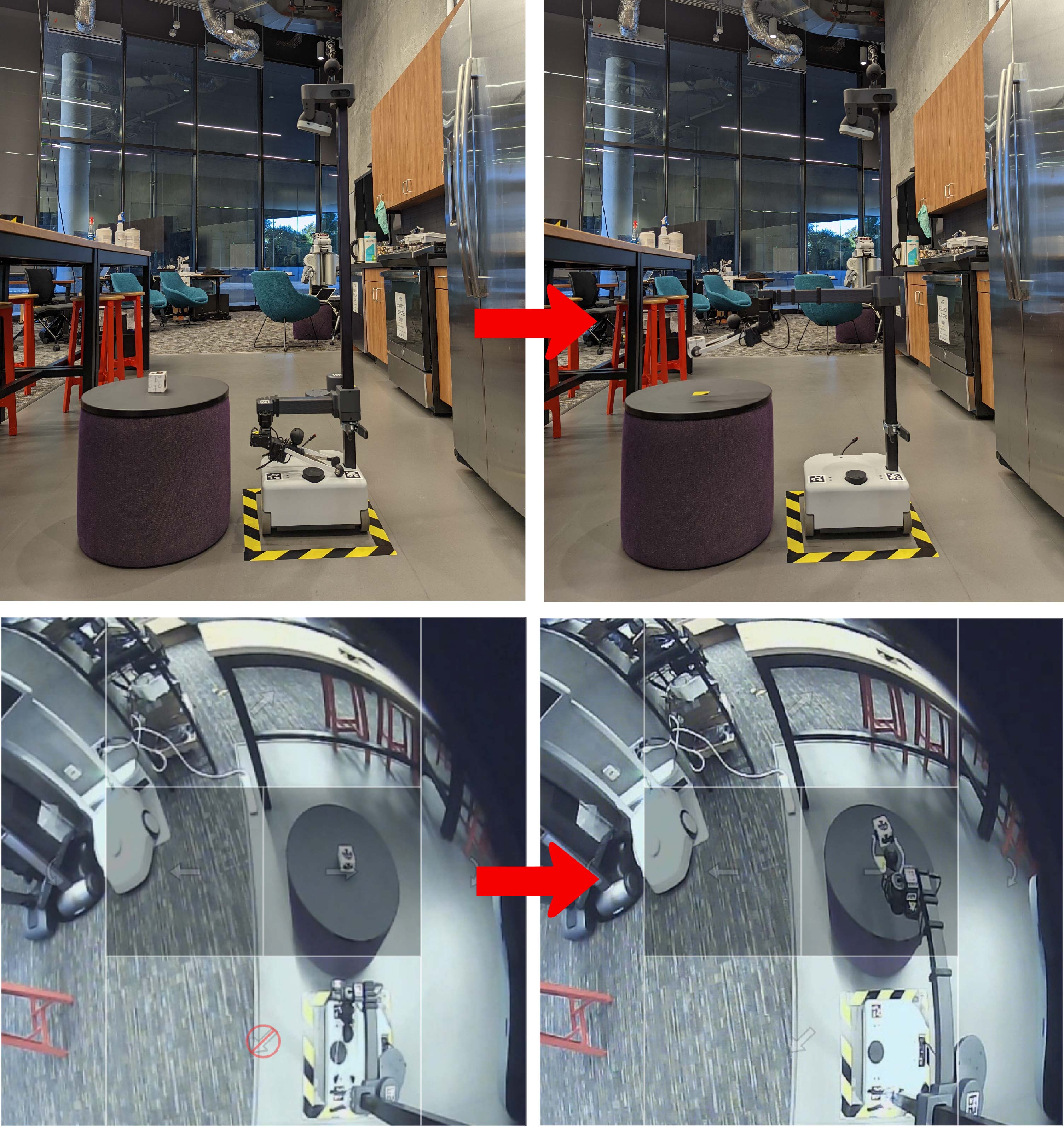}
         \caption{Task 2: Pick up cube on the table}
         \label{fig:task_2}
     \end{subfigure}
     \hfill
     \begin{subfigure}[b]{0.4025\textwidth}
         \centering
         \includegraphics[width=\textwidth]{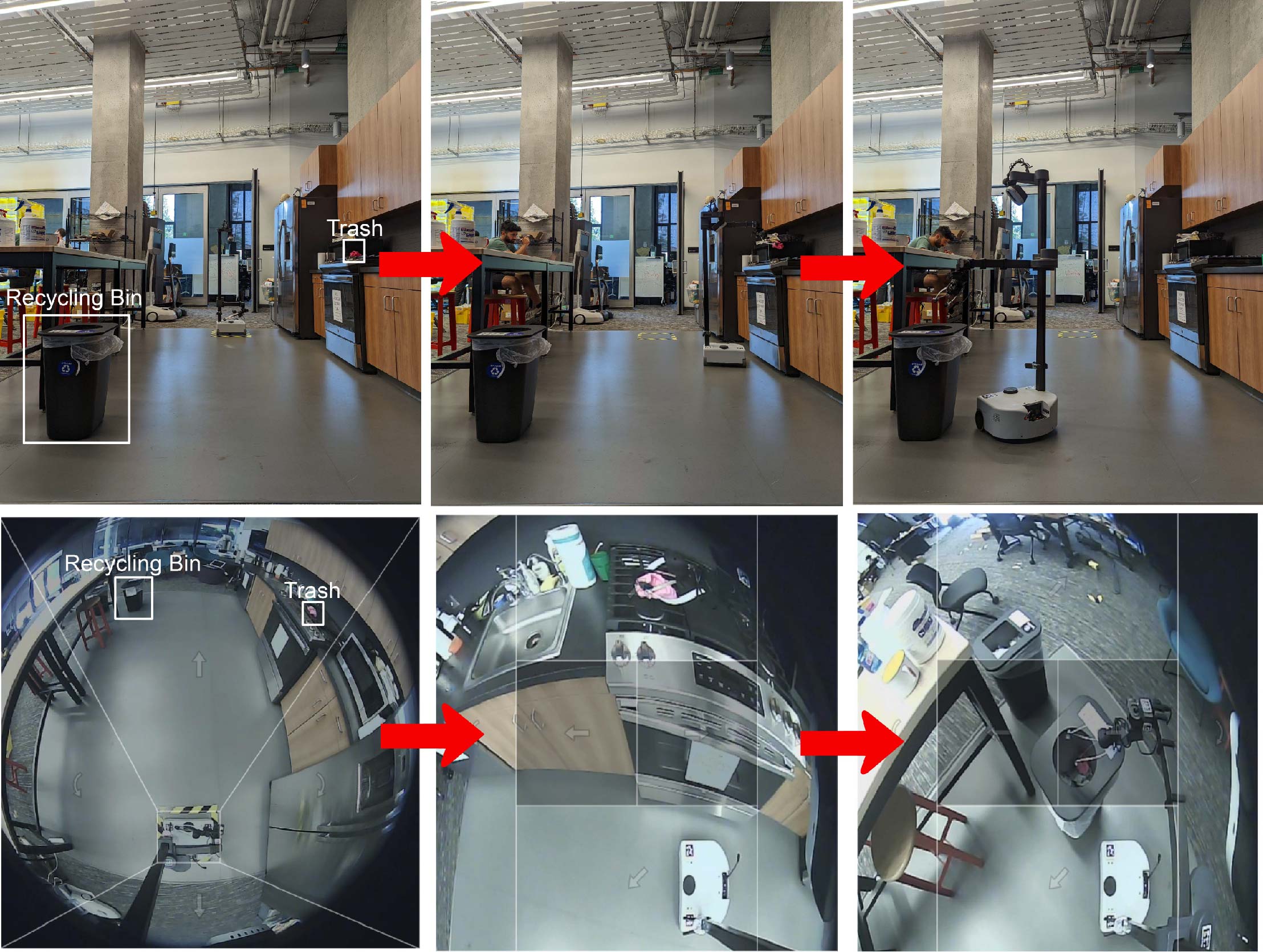}
         \caption{Task 3: Toss trash on the stove in the recycling bin}
         \label{fig:task_3}
     \end{subfigure}
    \caption{Overview of all tasks}
    \label{fig:tasks}
\end{figure*}

\section{Robot System}
\subsection{Hardware}
The Stretch RE1 mobile manipulator is developed by Hello Robot. Stretch has a telescoping arm that extends 50cm horizontally and is attached to a prismatic lift that reaches 110cm vertically. The arm has a 1 degree of freedom gripper attached to a rotational joint. The movement of the arm is orthogonal to the movement of the differential drive base. Stretch also has a Realsense camera attached to a pan-tilt head and two fixed fish-eye cameras: one with an overhead view of the base and arm and the other with a view of the gripper. 

Stretch's affordability, safety, and physical capabilities make it feasible to deploy long-term in novice users' homes. Stretch's software is open-source and is based on ROS1. It has a suite of autonomous features but our work focuses on remote tele-operation of the robot through a web interface. 

\subsection{Remote Tele-operation Interface Design}
The remote tele-operation interface (Fig.~\ref{fig:interface}) for Stretch has two distinct modes that can be toggled by switching tabs on the top left corner of the interface. Each mode has controls for controlling a different subset of the robot's actuators. The \nav mode controls the mobile base and the \manip mode controls the arm height, extension, and gripper. The \nav mode has two camera views: (1) a fixed overhead fish-eye camera view and (2) a overhead camera view with pan/tilt controls. The \manip mode has the same two camera views as the \nav mode but also has an additional fish-eye camera view from the gripper's perspective. Each mode has it's own subset of \textit{control displays} and \textit{action modes}.

\subsubsection{Action Overlay Control Display}
This control display has buttons overlaid on each camera view. The buttons control different actuators on the robot. The \nav mode has two translation and two rotation actions. The \manip mode has two buttons to control each of the following degrees of freedom: the arm's height, the arm's extension, gripper rotation in/out, open/close the gripper, and translation for the mobile base (for a total of 10 buttons). When the cursor hovers over a button, an icon overlaid indicates the action and tooltip text appears with explanation. Additionally, in the \manip mode, the icon turns red when when robot's arm or gripper is in collision with an object and a red stop sign appears over the icon when the arm and gripper have reach their respective joint limits. The user can control the speed of the robot by selecting from five preset speeds. The button outline turns red while the robot executes the corresponding action. 

\subsubsection{Predictive Display Control Display}
This control display overlays a trajectory on the fixed overhead fish-eye view of the robot's base and is only applicable in the \nav mode. The length and curve of the trajectory affect the speed and heading over the robot respectively. The longer the trajectory, the faster the robot will move and the shorter the trajectory the slower the robot will move. If the user presses anywhere behind the base the robot will move at a fixed speed backwards. If the user presses on the left side of the base the robot will rotate to the left and will rotate to the right if the user presses on the right side of the base. The trajectory turns red when the robot is moving. 

\subsection{Action Modes}
All action modes are applicable to both control displays. 
\begin{itemize}
    \item \textbf{Step Actions}: The robot moves for two seconds when user presses the button or trajectory. The distance the robot moves is determined by the speed. 
    \item \textbf{Press-Release}: The robot moves when the user presses and holds the button or trajectory and stops when the they release. 
    \item \textbf{Click-Click}: The robot moves when the user clicks the button or trajectory and stops when they click again. 
\end{itemize}

\section{User Study Design}

\subsection{Environment \& Tasks}
The study was conducted in a kitchen settings with a working area of roughly 2.15x4 meters. The tasks involve driving the robot to a specific position and orientation, picking up a cube, and recycling trash. Tasks involve a combination of observing the environment, navigating, manipulation and collision avoidance.

\textit{Task 1} - The user drives the robot from a starting position into a square in front of the fridge. They must orient the robot to face the fridge (Fig.~\ref{fig:tasks}\subref{fig:task_1}).

\textit{Task 2} - The user must control the robot to pick a cube up off a table. The robot is positioned next to the table (Fig.~\ref{fig:tasks}\subref{fig:task_2}).

\textit{Task 3} - The user must drive the robot from the fridge to the stove, pick up a piece of trash on the stove, drive to the recycling bin and drop the trash in the bin (Fig.~\ref{fig:tasks}\subref{fig:task_3}). 

\subsection{Procedure}
Participants join a video conferencing call through Zoom with screen sharing capabilities. They then log into the web interface for controlling the robot. Participants were located all around the U.S. and were not physically present. The user begins by watching an overview video of how the robot and interface works. In the first phase of the study, the user explores how to use the action overlay control display in the navigation mode. They watch video tutorials on how to use each of the action modes. After each video they have a chance to become comfortable with the controls. They then complete task 1 with the default settings and customized settings. They customize their settings by selecting their preferred action mode in the settings menu. The default setting is the step actions mode, which was the original action mode provided by the interface developed by Hello Robot. 

In the next phase of the study, the user explores that predictive display mode inspired by the Beam tele-presence robot interface. Similar to the first phase, they user watches video tutorial on how to use each of the action modes in this control display. After each video they have a chance to become comfortable with the controls. They then complete task 1 with the default and customized settings. The default setting is the press-release mode. This is the original action mode provided by the Beam's interface. Task 1 is considered a success when the user successfully drives the robot to the goal region and has it face the fridge. 

Next, the user explores the manipulation mode. They watch video tutorials on how to use each of the action modes and have a chance to become comfortable with the controls. They then complete task 2 with the default and customized settings. The default setting is the step actions mode. Task 2 is considered a success when the robot picks the cube up off the table. 

In the last phase of the study the user completes task 3 using both the default and customized settings for both the navigation and manipulation mode. Task 3 is considered a success when the trash is dropped in the recycling bin. The default settings for the action overlay and predictive display control displays are step actions and press-release respectively. The default setting for the manipulation mode is step actions. After selecting their preferred settings, the user fills out a questionnaire on the customization process. Additionally, the user fills out a questionnaire after completing the task with the default settings and then again with the customized settings. After completing the task twice, the user fills out a series of questionnaire about their experience, provides suggestions and recommendation and demographic information. Note, we counterbalance the order in which the task is completed with default and customized settings. 
\begin{figure}
\centering
  \includegraphics[width=\columnwidth]{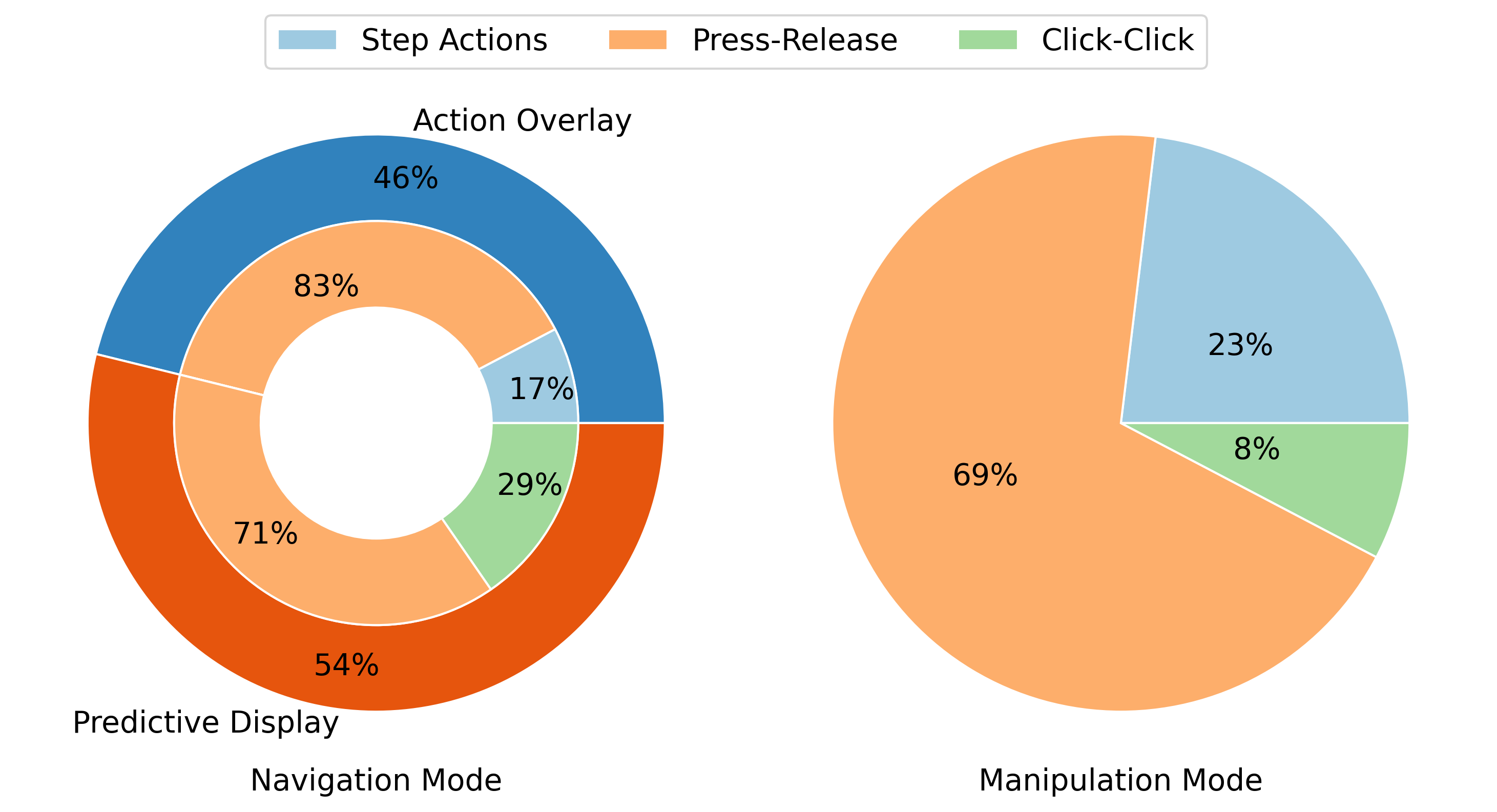}
  \caption{Users without motor impairments settings preferences for Task 3.}
  \label{fig:preferences_no_impairments}
\end{figure}

\subsection{Measurements}


During the study the users share their screen and we record both the users' verbally expressed thoughts and use of the interface. For each task we record the number of clicks, task completion time, whether or not the task was successfully completed, number and type of errors, including the user missing when grabbing an object or dropping an object. 

After completing each attempt of task 3, we asked users to state their agreement with a series of statements on a 5-point likert scale on usability, accessibility, efficiency of the interface and their satisfaction with the settings. Additionally, the user answered a series of open-ended questions on whether they found the robot useful, if they would use the robot in their homes, any modifications they would need to make to their home to use it and recommendations for improving the interface. 

\section{Findings}

\begin{figure}
\centering
  \includegraphics[width=\columnwidth]{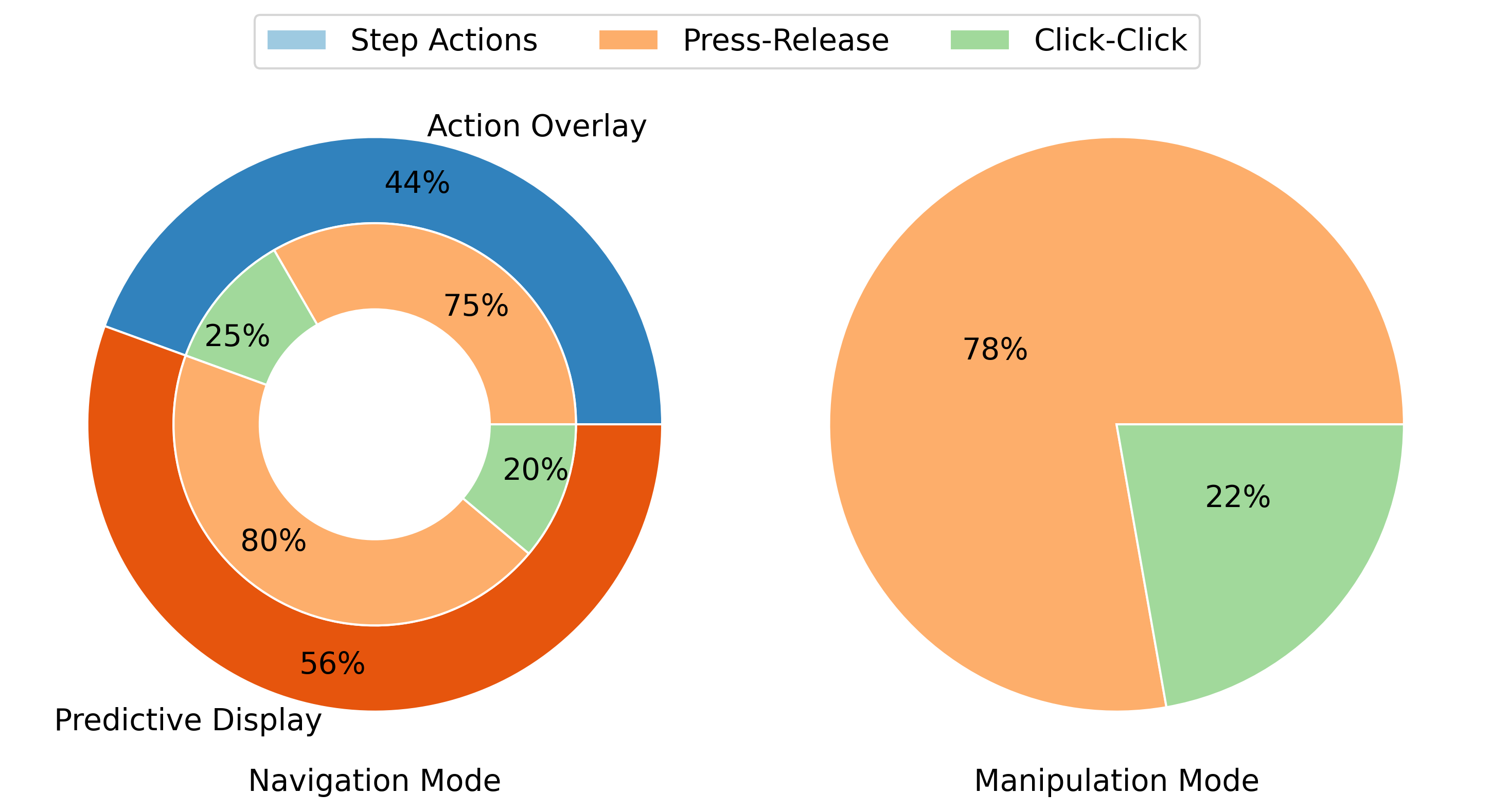}
  \caption{Users with motor impairments settings preferences for Task 3.}
  \label{fig:preferences_impairments}
\end{figure}

\subsection{Study 1: Users without motor impairments}
Our study was completed by 13 individuals from the general population (6 Male, 6 Female, 1 Other) with ages ranging from 20-55 (M=26, SD=9). We asked participants to rate their proficiency with technology on a 7-point Likert scale. The average rating was 5.31 with a standard deviation of 1.97. The study took 90 minutes and participants were compensated with a \$50 amazon gift card. We present our findings summarizing user setting preferences, task performance (success and efficiency), and then describe the tele-operation interface usage.

\subsubsection{Setting Preferences}
The preferred settings by participants for Task 3 are shown in Fig.~\ref{fig:preferences_no_impairments}. In the navigation mode, 46\% of participants chose the action overlay control display and 54\% of participants chose the predictive display control display. Participants who chose predictive display liked the simplicity in comparison to the action overlay mode: \textit{``Obviously, the predictive display is very nice, because it gets rid of buttons" (M, 25).} We asked participants to rate their proficiency with technology on a 7-point Likert scale. Majority of participants who chose action overlay rated themselves lower (M=3.83, SD=1.86) than participants who chose predictive display (M=6.57, SD=0.49). 


Overall, the press-release action mode was largely preferred in both the navigation and manipulation mode: \textit{``I like [press-release] mode better. In the [step-actions] mode it was to touch, stop, touch, stop"} (M, 29). 
However, there is no subset of settings that is a clear ``winner" as there is a spread across preferred control display and action mode. Note that 7.69\% of participants, across the different modes, chose their customized setting to be exactly the same as the default setting.

\subsubsection{Task Success}
All participants successfully completed Task 1 and 2 with both default and customized settings. All participants successfully completed Task 3 with the default settings and 11 participants successfully completed Task 3 with the customized settings.

Despite the high success rate, we observed errors during Task 2 and 3 (Table~\ref{table:errors}). The average number of errors for Task 2 was higher when participants used their customized settings. We noticed most errors occurred when users selected either press-release or click-click as their preferred mode. For Task 3 the average number of errors was lower when users used their customized settings. We did not see a correlation between proficiency with technology and number of errors.


\subsubsection{Task Performance}
We show the time taken and number of clicks across participants when using customized and default settings for all tasks in Fig.~\ref{fig:time-click-plots}. 

\begin{itemize}
    \item Task 1 - Action Overlay: Majority of the participants had fewer clicks and faster task completion time when using their customized settings.
    \item Task 1 - Predictive Display: Majority of the participants had fewer clicks and faster task completion time when using their customized settings.
    \item Task 2: Majority of participants had faster task completion time when using the default settings (step actions) irrespective of ordering. There is no clear trend for the number of clicks.
    \item Task 3 - Manipulation: Majority of participants spent less time in the manipulation mode when completing the task a second time irrespective of ordering. Participants generally had fewer clicks when they completed the task with their customized settings.
    \item Task 3 - Navigation: Majority of participants completed the task faster when using their customized settings but had fewer clicks their second time completing the task irrespective of ordering.
\end{itemize} 

\begin{figure}
\centering
  \includegraphics[width=\columnwidth]{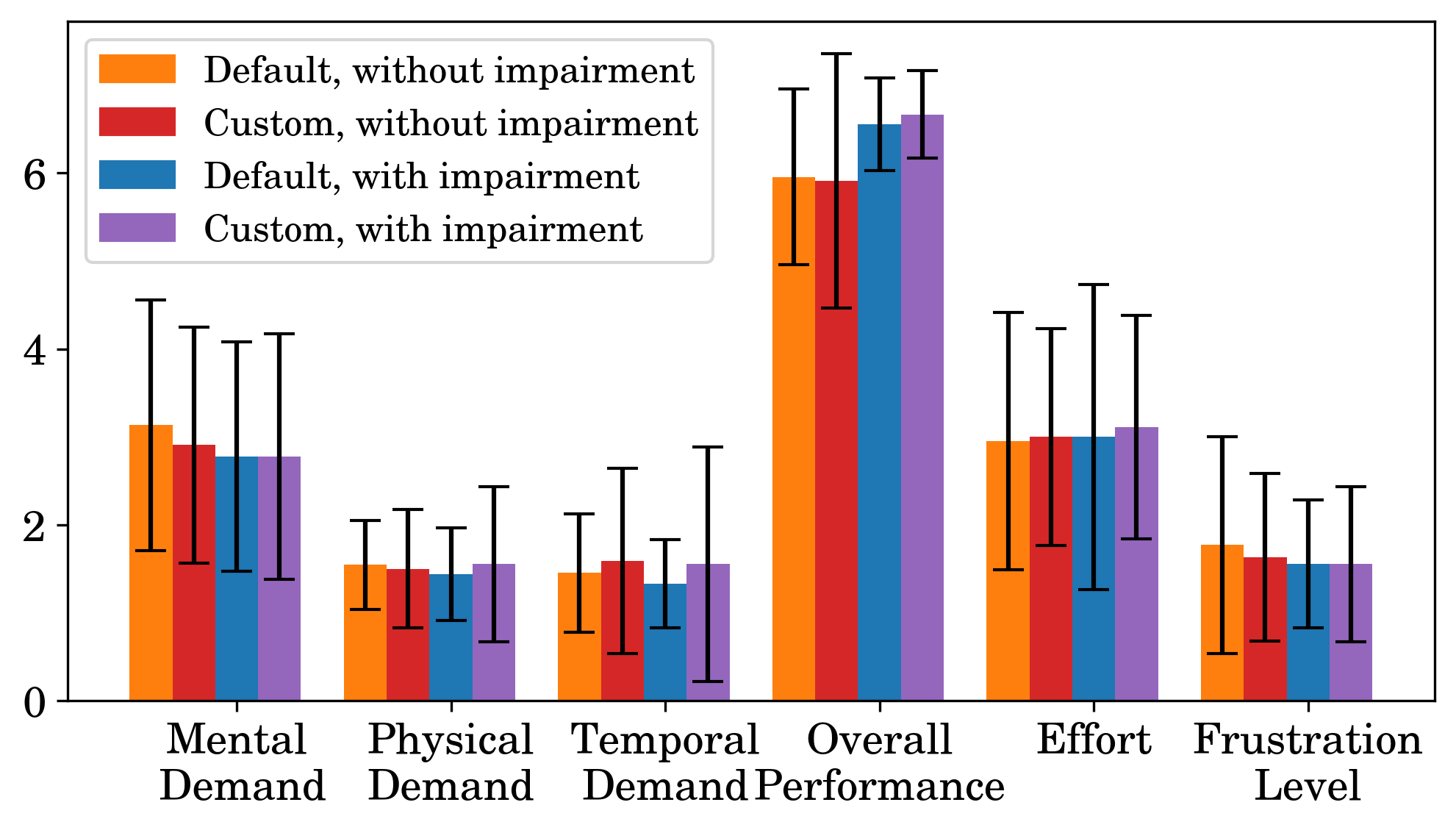}
  \caption{Task 3 workload for users with and without motor impairments.}
  \label{fig:tlx}
\end{figure}

Overall, participants had faster task completion time when using customized settings for navigation but had faster task completion time when doing the tasks a second time for the manipulation mode regardless of whether they used custom or default settings first. The manipulation mode is more difficult to use than the navigation mode as there are more buttons and degrees-of-freedom to control. One participant even found the manipulation mode to be overwhelming: \textit{``Having so many buttons makes me nervous"}. This possibly resulted in a learning curve irrespective of the task setting order for the manipulation mode.

\begingroup
\renewcommand{\arraystretch}{1.15}
\begin{table}
    \begin{center}
        \begin{tabular}{ c c c c c c c }
             \hline
             Task & Impairment & Setting Type & Avg & SD & Min & Max \\ 
             \hline
             1 & Yes & Default & 0 & 0 & 0 & 0 \\
             1 & Yes & Custom & 0 & 0 & 0 & 0 \\
             2 & Yes & Default & 0.5 & 0.92 & 0 & 3 \\ 
             2 & Yes & Custom & 0.2 & 0.4 & 0 & 2 \\
             3 & Yes & Default & 0 & 0 & 0 & 0 \\
             3 & Yes & Custom & 0.3 & 0.64 & 0 & 2 \\
             1 & No & Default & 0 & 0 & 0 & 0 \\
             1 & No & Custom & 0 & 0 & 0 & 0 \\
             2 & No & Default & 0.46 & 0.88 & 0 & 3 \\ 
             2 & No & Custom & 1 & 1.68 & 0 & 5 \\
             3 & No & Default & 0.46 & 0.88 & 0 & 3 \\
             3 & No & Custom & 0.31 & 0.48 & 0 & 1
        \end{tabular}
    \end{center}
    \caption{Number of errors across each task for participants with motor impairments}
    \label{table:errors}
\end{table}
\endgroup 

\begin{table*}
    \begin{center}
        \begin{tabular}{ c c l p{2.75in} p{1.5in} }
             \hline
             ID & Age & Source of Motor Impairment & Motor Impairments & Input Device\\ 
             \hline
             1 & 46 & C4 SCI & Paralysis in arms and legs & Head array \\
             2 & 21 & C5/6 SCI & Paralyzed from the chest down; no tricep or finger function & Trackpad and trackball mouse\\
             3 & 33 & CMT Type 2A & Paralyzed from the waist down; limited mobility in arms and hands & Standard computer mouse \\
             4 & 30 & C3 SCI & Paralysis in the arms, trunk and legs & GlassOuse \\
             5 & 27 & C5/6 SCI & Paralyzed form the chest down; no tricep or finger function & Trackball mouse \\
             6 & 31 & Transverse Myelitis & Paralyzed from the neck down; little hand movement & Drawing tablet with stylus operated with mouth \\ 
             7 & 22 & C5 SCI & Paralyzed from chest down; no tricep or finger function & Trackpad \\
             8 & 33 & C4/5 SCI & Paralyzed from the chest down; no tricep or finger function & Trackpad \\
             9 & 21 & C5/6 SCI & Paralyzed form the chest down; no tricep or finger function & Stylus \\
             10 & 27 & C4 SCI & Paralysis in arms and legs; has wrist mobility but not finger function & QuadJoy
        \end{tabular}
    \end{center}
    \caption{Demographic information of participants with motor impairments}
    \label{table:demographic}
\end{table*}

\subsubsection{Task Workload and Subject Evaluation}
\begin{figure*}[t!]
\centering
  \includegraphics[width=\textwidth]{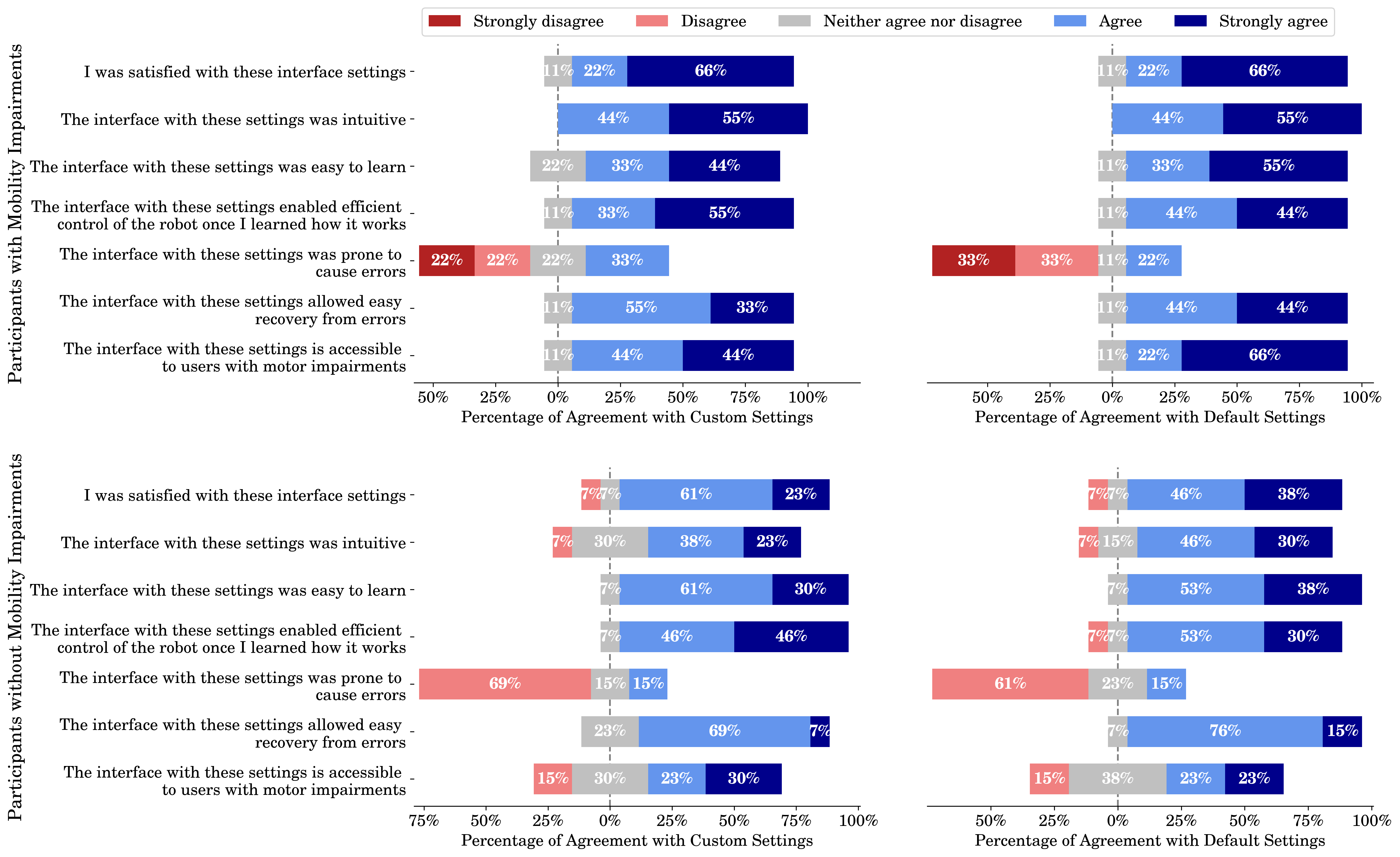}
  \caption{Participant agreement with statements about interface intuitiveness, efficiency and accessibility.}
  \label{fig:statements}
\end{figure*}

The task load index was assessed after the completion of Task 3 with both customized and default settings and the averages are shown in Fig.~\ref{fig:tlx}. On average, the mental demand, physical demand and frustration level ratings were slightly higher with the customized settings. Interface intuitiveness, learnability, efficiency, error recovery, accessibility to participants, and satisfaction with interface settings rated similarly between the default and customized settings~(Fig. \ref{fig:statements}).

\subsubsection{Utility of Robot}
All participants found the robot to be useful. Some said that the robot would not be useful in their own lives but could be useful to someone with motor impairments. Other participants said that the robot could be useful to complete tasks when they are not physically present or in hard to reach places such as overhead cabinets or shelves. One participant said the robot would be useful if they were sick and unable to get out of bed. They would use the robot to fetch things for them in this situation. 

\subsection{Study 2: Users with motor impairments}
Next, our study was completed by people with motor impairments who are our representative target population. The study setup and procedure is identical to the first study. We had 10 participants(4 Male, 6 Female) with varying levels of motor limitations (Table~\ref{table:demographic}) and ages ranging from 21-46 (M=30, SD=7.5). We asked participants to rate their proficiency with technology on a 7-point Likert scale. The average rating was 6 with a standard deviation of 0.87. The study took 90 minutes and participants were compensated with a \$100 amazon gift card. 

\subsubsection{Setting Preferences}
The preferred settings by participants for Task 3 are shown in Fig.~\ref{fig:preferences_impairments}. In the navigation mode, 44\% of participants chose the action overlay control display and 56\% chose the predictive display control display. 
Majority of participants preferred the press-release mode over other action modes for both the action overlay and predictive display control display: \textit{``The [press-release mode] is way easier than the [step actions mode]"} \textbf{(P3)}. Participants also noted that they liked the ability to take small steps within the press-release mode (as in the step actions mode), hence allowing for both continuous and step-wise control: \textit{``I like the [press-release mode] because you could just click, click, click for step-wise movement"} \textbf{(P8)}. None of the participants preferred the step-actions mode because of the fatigue caused by repetitive clicking.

Overall, similar to the preferences of people without motor impairments, the press-release action mode was largely preferred in both the navigation and manipulation mode. Again, there is no subset of settings that is a clear ``winner" as there is a spread across preferred control display and action mode. 

\begin{figure*}
\centering
  \includegraphics[width=\textwidth]{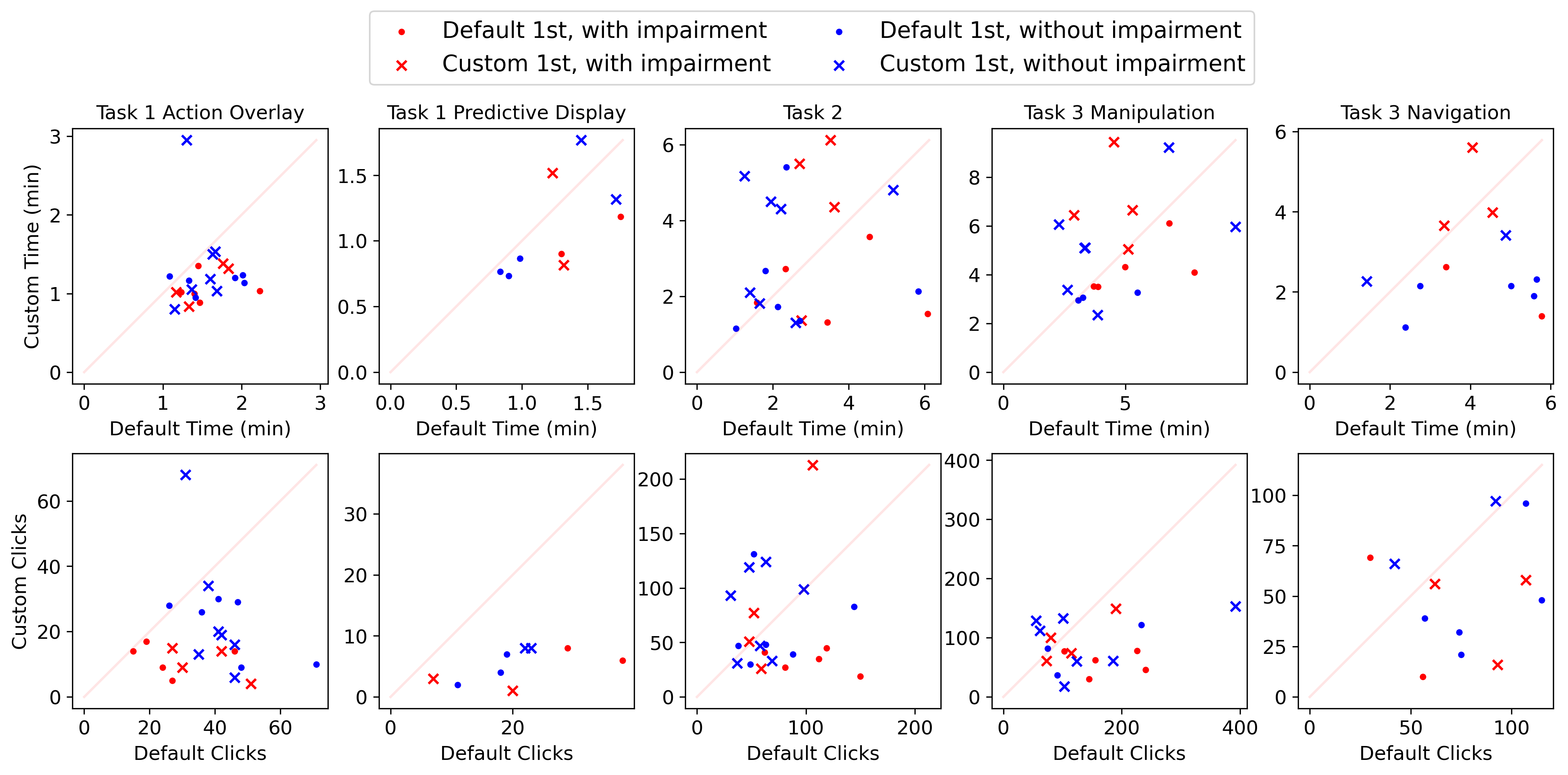}
  \caption{(Top) The time taken when using default settings versus customized settings. (Bottom) The number of clicks when using default settings versus customized settings. Points under the line show that users had fewer clicks or completed the task faster with the customized settings. Points above the line show that users had fewer clicks or completed the task faster with default settings. We only plot points for users that chose settings different than the default settings.}
  \label{fig:time-click-plots}
\end{figure*}

\subsubsection{Task Success}
All participants successfully completed all three tasks. We observed errors in Task 2 with the default and customized settings and Task 3 with the customized settings (Fig.~\ref{table:errors}). 
With the default setting in Task 2 (i.e., step actions), we noticed that some participants had difficulty estimating how far the arm would move based on their speed setting. This caused the robot to overshoot when reaching for the cube and collide with the table. One participant missed grabbing the cube twice in Task 2 with the customized settings and 2 participants missed grabbing the trash in Task 3 with the customized settings as they had trouble with depth perception. Overall, the number of errors was low and participants recovered from errors and eventually succeeded in completing the tasks.  

\subsubsection{Task Performance}
We show the time taken and the number of clicks across participants when using customized and default settings for Task 3 in Fig.~\ref{fig:time-click-plots}. \textbf{P1} is not included in this plot as they were not able to complete the task with the default settings for the predictive display control display (press-release). Their head array was not capable of doing the press and hold cursor action. 

\begin{itemize}
    \item Task 1 - Action Overlay: All participants had faster task completion time and fewer clicks when using their customized settings.
    \item Task 1 - Predictive Display: Majority of the participants completed the tasks faster when using their customized settings and all participants had fewer clicks when using their customized settings irrespective of ordering.
    \item Task 2: Majority of participants had faster task completion time and fewer clicks when completing the task a second time irrespective of interface settings.
    \item Task 3 - Manipulation: Majority of participants completed the task faster the second time irrespective of interface settings, but they had fewer clicks when using the customized settings.
    \item Task 3 - Navigation: Majority of participants had faster task completion time and fewer clicks when using the customized settings.
\end{itemize}

Overall, participants had faster task completion time when using customized settings for navigation but had faster task completion time when doing the tasks a second time for the manipulation mode irrespective of ordering. This is possibly a result of the manipulation mode being more difficult learn for the aforementioned reasons. P8 referred to this learning curve: \textit{``It's really fun. I think it's more a matter of you keep doing it and getting used to it. You're just figuring it out. It's like when you get a new phone, and you don't know where things are."} \textbf{(P8)}.

\subsubsection{Task Workload and Subjective Interface Evaluation}
The task load index was assessed after the completion of Task 3 with both customized and default settings. The averages are shown in Fig.~\ref{fig:tlx}. Overall, all TLX ratings were very similar between default and customized settings. Interface intuitiveness, learnability, efficiency, error recovery, accessibility to participants, and satisfaction with interface settings rated similarly between the default and customized settings~(Fig. \ref{fig:tlx}). Additionally, the ratings for all categories are higher than the ratings by the participants without motor impairments. The difference in ratings between users with and without motor impairments is possibly due to the direct impact this platform could have on the users' lives. 

\subsubsection{Utility of Robot}
All participants said that the robot is useful and that their homes could accommodate the robot. All participants said that they would use the robot to retrieve items around the household such as water (P6, P9), cooking utensils (P7), food (P7) and medical supplies (P5). Participants also said they would use it for tasks such as scratching their forehead (P6), unloading laundry (P3), putting groceries away (P10), and organization (P10). Participants with arm function and no leg function said that they would specifically use the robot to fetch them items that are beyond their reach and if they were in their bed instead of their wheelchair: \textit{``The last place I was staying at I literally, did not get out of bed for like a month, and this would have been nice to get my water out of my fridge."} \textbf{(P9)}. Participants with no arm or leg function said they would use it more frequently so that they would not need to ask anyone for help. Most participants did not find utility for the robot outside of their home, but P3, P8 and P10 said that they could use the robot when grocery shopping. 

We asked participants to rate their independence on a 7-point Likert scale (M=2.6, SD=1.02). We then asked participants to state their agreement with the statement ``Having the robot in my home will make me more independent" on a 7-point Likert scale (M=5.8, SD=1.6). Overall, participants believe the robot is useful and will make them more independent.

\section{Discussion \& Conclusion}
This paper explored customization of tele-operation interfaces for assisting individuals with severe motor limitations and potential caregiver. We believe that we would see greater benefits of customization if the robot was deployed long-term in someone's home (e.g.~\cite{nguyen2022hfes}). Nevertheless, this work confirms the utility of such a robot and the benefits of customization of tele-operation interfaces for both user groups.

\subsubsection{Settings Preferences}
User preferences in interface configurations varied. There was no single interface configuration that was more strongly preferred over another. Users with motor impairments did not choose the step actions mode due to the fatigue of repeated clicks. Some participants without motor impairments chose the step actions mode in the action overlay control display because they were worried that they would damage the robot in the continuous control modes. Additionally, P1 was not able to use the press-release mode with his head array. These findings confirm our hypotheses that there is no single interface configuration that satisfies all users' abilities and preferences (H1) and there are difference in preferences between participants with and without motor impairments (H2).

\subsubsection{Task Performance}
All users had faster task completion time and fewer clicks with the customized settings in the navigation mode but performed better the second time in the manipulation mode irrespective of which interface configuration they started with. This suggests that they was a learning curve which is possibly due the complexity of the interface controls in the manipulation mode. We believe if users had more time to familiarize themselves with the manipulation mode they would have performed better with the customized settings. Additionally, users did not have a practice task that combined both navigation and manipulation mode. We noticed a learning curve associated with using both modes to complete a task. Overall, the number of errors was very low, all participants successfully completed task 1 and 2 and all but two participants without motor impairments successfully completed task 3. These findings partially confirm our hypothesis (H3) that users' task completion time, number of errors and clicks will be lower when using their customized settings.


\subsubsection{Context Adaptation}
Several participants wanted to switch between different settings when completing task 3. For example, some participants wanted to use the step actions when trying to pick up or drop the trash in the recycling bin. Some participants who selected the press-release mode realized that they could use it like the step actions mode with shorter clicks. This suggests that settings preferences can vary depending on the context of the task and interfaces should allow for easy adaptation of settings to different contexts.

\addtolength{\textheight}{-12cm}   


\bibliographystyle{IEEEtran} 
\bibliography{IEEEfull.bib}

\end{document}